\documentclass[sn-mathphys,Numbered]{sn-jnl}% Math and Physical Sciences Reference Style
\usepackage{mathptmx}
\usepackage[table,xcdraw]{xcolor}

\usepackage{graphicx}%
\usepackage{multirow}%
\usepackage{amsmath,amssymb,amsfonts}%
\usepackage{amsthm}%
\usepackage{mathrsfs}%
\usepackage[title]{appendix}%
\usepackage{xcolor}%
\usepackage{textcomp}%
\usepackage{manyfoot}%
\usepackage{booktabs}%
\usepackage{algorithm}%
\usepackage{algorithmicx}%
\usepackage{algpseudocode}%
\usepackage{listings}%
\usepackage{ragged2e}
\usepackage{adjustbox}
\usepackage{comment}
\newcommand{\del}[1] {{\color{teal} #1}}
\begin{document}
%conflation of
\title[Article Title]{%Requirements Modeling for Compliance and Conformity of Trustworthy AI 
An Open Knowledge Graph-Based Approach for Mapping Concepts and Requirements between the EU AI Act and International Standards} %  


\author*[1]{\fnm{Julio} \sur{Hernandez}}\email{julio.hernandez@adaptcentre.ie}

\author[1]{\fnm{Delaram} \sur{Golpayegani}}\email{delaram.golpayegani@adaptcentre.ie}

\author[1]{\fnm{Dave} \sur{Lewis}}\email{dave.lewis@adaptcentre.ie}

\affil*[1]{\orgdiv{School of Computer Science and Statistics}, %\orgname{ADAPT Centre, Trinity College Dublin}, \orgaddress{\street{College Green}, \city{Dublin}, \postcode{D02PN40}, \state{Dublin}, \country{Ireland}}}

\orgname{ADAPT Centre, Trinity College Dublin (TCD)}, \orgaddress{\state{Dublin}, \country{Ireland}}}

%\affil[2]{\orgdiv{Department}, \orgname{Organization}, \orgaddress{\street{Street}, \city{City}, \postcode{10587}, \state{State}, \country{Country}}}

%\affil[3]{\orgdiv{Department}, \orgname{Organization}, \orgaddress{\street{Street}, \city{City}, \postcode{610101}, \state{State}, \country{Country}}}


\abstract{The many initiatives on trustworthy AI result in a confusing and multipolar landscape that organizations operating within the fluid and complex international value chains must navigate in pursuing trustworthy AI. The EU's AI Act will now shift the focus of such organizations toward conformance with the technical requirements for regulatory compliance, for which the Act relies on \emph{Harmonized Standards}. Though a high-level mapping to the Act's requirements will be part of such harmonization, determining the degree to which standards conformity delivers regulatory compliance with the AI Act remains a complex challenge. Variance and gaps in the definitions of concepts and how they are used in requirements between the Act and harmonized standards may impact the consistency of compliance claims across organizations, sectors, and applications. This may present regulatory uncertainty, especially for SMEs and public sector bodies relying on standards conformance rather than proprietary equivalents for developing and deploying compliant high-risk AI systems. To address this challenge, this paper offers a simple and repeatable mechanism for mapping the terms and requirements relevant to normative statements in regulations and standards, e.g., AI Act and ISO management system standards, texts into open knowledge graphs. This representation is used to assess the adequacy of standards conformance to regulatory compliance and thereby provide a basis for identifying areas where further technical consensus development in trustworthy AI value chains is required to achieve regulatory compliance.}

\maketitle
\textcolor{red}{This work was presented at the 9th International Symposium on Language \& Knowledge Engineering (LKE 2024) Dublin, Ireland, 4 - 6 June, 2024.}


\section{Introduction}\label{introduction}

The global interest in AI's ethical and societal risks has grown rapidly in recent years~\cite{yigitcanlar2020contributions, nasim2022artificial, floridi2021ethical, o2020bias}. In the primary wave of trustworthy AI initiatives, guidelines typically are presented as structured statements of principles that organizations can adopt to demonstrate some degree of trustworthiness in their development and use of AI technology. With the increasing number of AI incidents, it became evident for policymakers and public authorities that there is a wide range of applications through which AI negatively impacts people's lives that are developed and deployed with little external oversight~\cite{beckman2022artificial, europeanAIStrategy}. Consequently, several jurisdictions are now developing AI legislation to introduce oversight over the development and use of AI, ensuring individuals, groups, and society are protected from its potential harms. 

 With its political agreement on the AI Act~\cite{aiAct} being reached at the end of 2023, the European Union (EU) has become a pioneer in AI regulation. The AI Act specifies a tiered risk system, where some applications of AI are \emph{prohibited}, and others are identified as a sufficiently low risk that only consumer labels or voluntary codes of practice are required. However, the focus of regulatory oversight and compliance information exchange lies between these tiers where high-risk AI systems are defined. The AI Act identifies high-risk AI application areas and requires that their development and deployment demonstrate conformance to risk and quality management measures in order to comply with the regulation. These measures follow the regulatory mechanism, called the New Legislative Framework (NLF)\footnote{\url{https://single-market-economy.ec.europa.eu/single-market/goods/new-legislative-framework_en}}, that is already established by the EU to provide a single health and safety regulatory framework for products across the European Single Market. The AI Act extends this mechanism to products and services containing AI and extends the scope of protection beyond health and safety to include the protection of all fundamental rights and the environment. In this way, the legislation aims to build public trust in AI while encouraging innovation in AI value chains by normalizing regulatory oversight. 

While the EU has separately provided guidelines for developing trustworthy AI\footnote{\url{https://digital-strategy.ec.europa.eu/en/library/ethics-guidelines-trustworthy-ai}}, these do not form part of the AI Act, given their principle-based representation. Instead, the detailed rulemaking on implementing the Act, including how AI risks are assessed, managed, and monitored, are delegated to technical standards. These can be in the form of standards that have been harmonized with the requirements of the AI Act by European Standardisation Organisations (ESO), namely the Comité Européen de Normalisation (CEN), Comité Européen de Normalisation Electrotechnique (CENELEC), or the European Telecommunications Standards Institute (ETSI). In response to the European Commission's draft standardization request~\footnote{\url{https://single-market-economy.ec.europa.eu/single-market/european-standards/standardisation-requests_en}}, relevant standards are already being addressed internationally by European standards development bodies such as ISO/IEC JTC 1 Subcommittee 42\footnote{\url{https://www.iso.org/committee/45020.html}} on AI (SC42). However, these development initiatives involve complex sets of interrelated standards, many of which are still under development~\cite{aiLandscape} and will evolve parallel to the AI Act and similar legislation being considered in other jurisdictions. 

With the forthcoming enforcement of the AI Act, one of the key challenges for high-risk AI providers and deployers is navigating in a sea of standards for addressing trustworthy AI requirements through regulatory compliance. In this, the lack of common terminology and detailed mapping of requirements adds to the complexity faced by providers and deployers. Any mappings between legal requirements for trustworthy AI and technical standards that enable conformance and certification functions that satisfy those requirements will require flexible, extensible, transparent, and auditable solutions to satisfy regulatory and organizational rules on governance process integrity. Open standards should be used, as far as possible, to increase third-party inspection and, therefore, confidence in the completeness and accuracy of mapping. In this paper, we take an approach based on Open Knowledge Graphs (OKG) specified using standards from the World Wide Web Consortium (W3C), which have been proven to be successful in promoting interoperability between approaches, satisfying the requirements of the EU General Data Protection Regulation (GDPR)~\cite{PANDIT2018262} and expressing high-risk information through an AI risk ontology based on the requirements of the AI Act and ISO 31000 series of standards~\cite{Golpayegani2022AIROAO}.

\section{Related Work}\label{related_work}

%semantic mapping of regulatory requirements to standards and TAI principles \cite{golpayegani2022comparison}

%mapping of AI Act's requirements to JTC 1/SC 42 standards (not semantic web) \cite{golpayegani2023high} and \cite{soler2023analysis}

%trustworthy AI ontologies \cite{Lewis21tai}
%\cite{amaral2020ontology}
There is some existing work addressing the challenges of implementing trustworthy AI requirements through utilizing OKG-based approaches.
%Some ontology approaches to map standards of trustworthiness have been proposed~\cite{golpayegani2022comparison,Lewis21tai,amaral2020ontology}. 
Amaral et al.~\cite{amaral2020ontology} combine the Reference Ontology for Trust (ROT) and the Non-Functional Requirements Ontology (NFRO) to characterize an ontology that captures trust requirements for software systems. Inspired by ISO/IEC JTC 1/SC 42 activities, Lewis et al.~\cite{Lewis21tai} propose a high-level ontology to map out the consistency and overlap of concepts from different AI standards, regulations, and policies. Golpayegani et al.~\cite{golpayegani2022comparison} use the aforementioned ontology to compare the semantic interoperability between ISO/IEC 42001 standard on AI management system, the EU trustworthy AI assessment list (ALTAI) and the EU AI Act. In this work, are map AI concepts and requirements from regulations and standards to develop a mechanism to compare, integrate, and relate the terminology used by these documents with the objective of regulatory compliance.

\section{AI Act Compliance Through Conformity with Standards}\label{AI_Act_Compliance}

This section presents an analysis of the AI Act and harmonized standards, followed by an analysis of legal compliance challenges with standards.

\subsection{The Interplay between the AI Act and Harmonized Standards}

Following the mechanisms established in the NLF, providers of high-risk AI applications need to demonstrate their compliance with the essential requirements of the AI Act through a conformity assessment process that is either self-certified or certified by a recognized authority, known as a \emph{notified body}. The conformity assessment process must address AI Act requirements related to risk management, data governance, and technical documentation under a quality management system for the product's compliance to be certified. This mechanism aligns well with the standards developed by the ISO Committee on Conformity Assessment (CASCO) through the ISO 17000 series of standards that guides the terminology, concepts, requirements, processes, and competencies regulators can use to establish certification rules. These standards are then complemented by standards that an organization can follow to implement the risk and quality systems compatible with CASCO-defined certification, known as Management System Standards (MSS). 

The ISO/IEC 42001\footnote{\url{https://www.iso.org/standard/81230.html}} is an AI MSS released by ISO/IEC JTC 1 SC42. Thus, this AI MSS and other SC42 standards reference form a strong candidate for adoption by the European Standardization Organization (ESO) in response to a harmonized standards request from the European Commission (EC) for the AI Act, given their alignment with an existing standardized conformance framework. 

\subsection{Challenges of Legal Compliance Using Harmonized Standards}
Several challenges remain when considering the vertical nature of the AI Act’s high-risk classification, the potentially complex value chains involved, and the international nature of AI innovation. \textbf{First,} the AI Act focuses its provisions for high-risk AI based on a specific set of applications, categorized into two groups: 
(i) AI systems that are products or safety components of the products already subject to the \emph{Union harmonization legislation}— a set of specific European product health and safety regulations, %e.g. in products 
such as \del{regulations on} machinery, toys, medical devices, agricultural vehicles, and rail systems.  
(ii) AI applications that are not yet regulated but are identified by the EC as presenting high risks to health, safety, or fundamental rights. However, the technical requirements for compliance with the AI Act and the potential harmonized standards from SC42 are horizontal, i.e., specified in terms that apply to any AI system. For instance, if we consider the risk of a voice recognition system misunderstanding the same utterance in different accents, the acceptable risk level when used in ambulance dispatch may involve different considerations from use in primary school student assessment. 

\textbf{Second,} many AI providers may already be undertaking some form of proprietary trustworthy AI risk assessment and quality process, e.g., The Microsoft Responsible AI Standard, v2\footnote{\url{https://blogs.microsoft.com/wp-content/uploads/prod/sites/5/2022/06/Microsoft-Responsible-AI-Standard-v2-General-Requirements-3.pdf}}. Such AI providers will need to undertake a mapping to assess whether the proprietary approach fully satisfies the requirements of the AI Act. They may also wish to establish a transition mapping from the proprietary standard to a relevant harmonized standard to reduce the cost of demonstrating compliance with the AI Act, which is estimated to be between 193,000€ to 330,000€\footnote{\url{https://www.cecimo.eu/wp-content/uploads/2022/10/CECIMO-Paper-on-the-Artificial-Intelligence-Act.pdf}}, and improving the potential for establishing such compliance, and thereby its trustworthy AI competencies to its customers and affected societal stakeholders more broadly. 

\textbf{Third}, there may be populations of AI providers that have invested in undertaking a trustworthy AI risk and quality assessment based on standards from national bodies, e.g., NIST\footnote{National Institute of Standards and Technology}, DIN\footnote{Deutsches Institut für Normung.}, BSI\footnote{British Standards Institution}, or other international standards, e.g. IEEE P7000~\cite{9410482}. Mapping between such standards and the AI Act’s harmonized standards may be important for AI providers to manage the cost of maintaining compliance with regulations in multiple jurisdictions. Providing such mappings could also support future equivalence agreements for trustworthy AI compliance between the EU and other jurisdictions regulating AI. 

The evolving nature of international standards for trustworthy AI exacerbates these requirements and compliance mapping challenges. There is a need for the harmonization request for the AI Act to be satisfied by European Standard Development Organizations (SDOs) that are not currently driving those standards and the proliferation of other proprietary, international, and national guidelines and standards for trustworthy AI. This paper presents an open approach to capturing requirements from different regulations and associated standards documents so that the sufficiency of the local management process and resulting artifact exchanges between value chain actors can be compared and compliance with different regulatory and policy requirements can be assessed and tracked from, e.g., auditors and notified bodies perspective.

\section{A Layered Approach for Semantic Modeling and Mapping of Trustworthy AI Requirements}\label{approach}

The challenges of mapping normative statements from regulations, such as the AI Act, against those in standards from different SDOs require cataloging the normative statements from these different source documents to mirror the granularity of authority and their revision cycles. This work analyzed the sections of the AI Act, specifically the compliance requirements for AI Providers for high-risk AI systems, and the terms and concepts defined by SC42 in foundational standards ISO/IEC 22989, as well as the template for ISO MSS, which forms the basis for the development of the AI MSS. %(\textbf{Figure~\ref{fig:layered_approach}}).

\begin{comment}
\begin{figure}
    \centering
    \fbox{\includegraphics[scale=0.29]{Figures/LayeredApproach.png}}
    \caption{Layered approach towards mapping requirements from separately developed documents}
\label{fig:layered_approach}
\end{figure}
\end{comment}

OKGs are grounded in the Resource Description Framework (RDF)~\cite{manola2004rdf}, which allows an unlimited knowledge graph of nodes and links to existing online resources on the web, thus lending themselves to third-party scrutiny. Nodes and associations in this knowledge graph are typed according to ontologies, also known as data vocabularies, that can be developed independently and published to a distinct namespace on the web. This highly decentralized approach aligns well to promote the participation of those generating standards, organizational policies, and regulations, as well as those interested in how these documents develop and map to each other. OKGs also offer predictable and controlled upgrade paths for expressing compliance rules as new regulations or regulatory guidance and case law emerge, allowing regulatory compliance for trustworthy AI to remain robust and cost-controlled amidst rapid evolution in the relevant regulation. 

In developing a semantic model for any specific domain, different levels of semantic commitment can be employed to express semantic relationships between possible information elements. The Web Ontology Language (OWL)~\cite{hitzler2009owl} allows information elements to be modeled as classes or instances, like object-oriented software engineering models. OWL classes can be structured hierarchically so that one class can be declared a subclass of another. Properties can be declared between classes and literal types that allow facts or axioms about the world to be asserted and inferred. 

However, trustworthy AI is a domain with a wide range of competing conceptual models but a relative paucity of concrete instances where trustworthy characteristics have been modeled, tested, and subject to third-party scrutiny. It is, therefore, more appropriate to capture some structure of knowledge without a full understanding of the instances that define the conceptual classes, the relationships between them, and the nature of any hierarchical structures, or we may not necessarily have the goal of checking data model consistency. In such scenarios, the Simple Knowledge Organization System (SKOS)~\cite{isaac2009skos} can organize concepts into concept sets and establish hierarchical relationships useful for building taxonomies. In SKOS, hierarchical associations are defined as a ‘narrower’ or ‘broader’ relationship between concepts, which makes no semantic commitment about these concepts being classes of instances and, therefore, makes no claims about the relationships of instances. The existence of concept relationships that do not have a hierarchical characteristic can also be captured by a ‘related’ association between those concepts. SKOS concepts and their associations can be grouped into concept sets representing the consensus developed on a domain by a group at a particular time. 

\section{TAIR: Trustworthy AI Requirements Ontology}\label{tair}

This section introduces some challenges in mapping normative statements from regulations and standards. Additionally, it presents the TAIR ontology as a semantic approach to map concepts and requirements from regulations and standards. Finally, the TAIR ontology evaluation is presented, considering the best practices for detecting errors in ontology design.   

\subsection{Conceptual Requirements Capture from AI Act and Prospective Harmonized Standards}

\begin{figure}[t]
    %\sidecaption
    \centering
    \fbox{\includegraphics[scale=0.25]{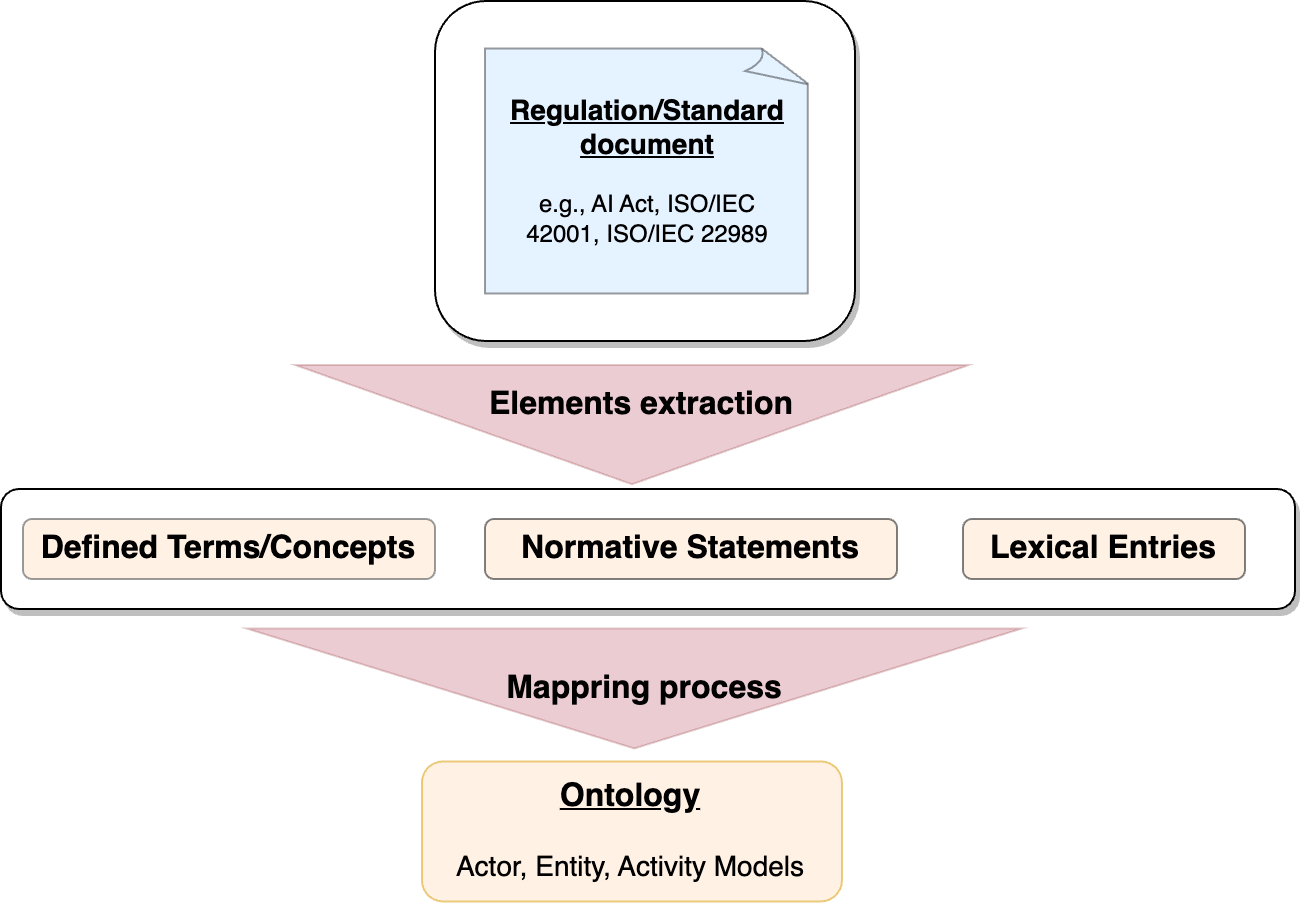}}
    \caption{Mapping concepts from regulations and standards}
    \label{fig:mapping_concepts}
\end{figure}

We aim to enable the capture of terms and concepts related to regulatory requirements and standards to which organizations in the AI value chain can conform to demonstrate their compliance with their regulatory obligations. The approach specifically aims to enable the interlinking of requirements between regulatory text and texts specifying such international standards, thereby checking the extent to which prospective harmonized standards requirements will deliver regulatory compliance. This requires an analysis of the normative scope of requirements of both the relevant compliance clauses of the AI Act and the AI MSS template (\textbf{Figure~\ref{fig:mapping_concepts}}).

Our semantic modeling leverages the core commonality of the harmonized structure for MSS~\cite{ISODIRECTIVES_PART1} to provide a minimal and reusable approach, determining the extent to which the requirements present in normative statements specified in a regulatory text for trustworthy AI are satisfied by normative statements in technical standards documents used in conformance, specifically those stemming from AI MSS. This is taken as a specific assessment of the more general goal to assess whether this approach allows machine-readable mapping for specific proposed trustworthy AI guidelines or standards to be mapped against requirements of specific regulatory text. The target forms of mapping consider: 

\begin{itemize}
    \item {Whether all captured regulatory requirements are addressed by the available management system or other technical requirements.}
    \item {Whether regulatory requirements have mappings to specific technical activities or entities/artifacts defined in the technical standards}
    \item {Whether some requirement mapping is partial in that they use a different definition of concepts or different levels of normative strictness, i.e., the requirement (must/shall) compares to a recommendation (should), permission (may), or possibility (can).}
    \item {If there are terms in the regulatory requirements for which mapping to technical standard requirements, activities, or entities cannot be fully determined.}
\end{itemize}

\textbf{Terms extraction.} The term extraction and mapping process first involves extracting explicitly defined terms as SKOS concepts. The structure of terminological lists (for example, subsection in the terminology section of ISO/IEC standards), the text of the definitions, and cross-references between these are used to capture taxonomical structures, using the SKOS ‘narrower’, ‘broader’, and ‘related’ relationships. 

\textbf{Requirements extraction.} Normative clauses of the source documents are converted to atomic normative requirements\footnote{A specific irreducible requirement involving named actors, activities, or entities}.  

\textbf{Lexical entries extraction.} Where the requirement or required situation from an individual normative statement conceptualization does not correspond to a term from that same source document, the terms used are captured as a lexical entry, indicating that it is a concept that may require further definition to support future compliance checking. Therefore, lexical entries are candidates for alignment with definitions from another document, e.g., from another referenced legislative document or technical standard. 

The use of SKOS concepts set for terms and approaches that aim to allow formal expression of a requirement that can be subject to deontic reasoning as part of a requirement management process. Instead, Our approach focuses on facilitating the mapping between separate developed sets of definitions and compliance standards in a flexible, extensible, and repeatable manner appropriate to the evolving trustworthy AI landscape where compliance rules and associated standards are still under development.

Next, we present the semantic web ontology used to conceptualize and map this term and requirements. 

\subsection{TAIR Overview}

This section presents the key elements of the TAIR ontology and the requirements and concepts of the semantic mappings process. 

\subsubsection{Requirements and Concepts Modelling}
\begin{figure}[t]
    %\sidecaption
    \centering
    \fbox{\includegraphics[scale=0.25]{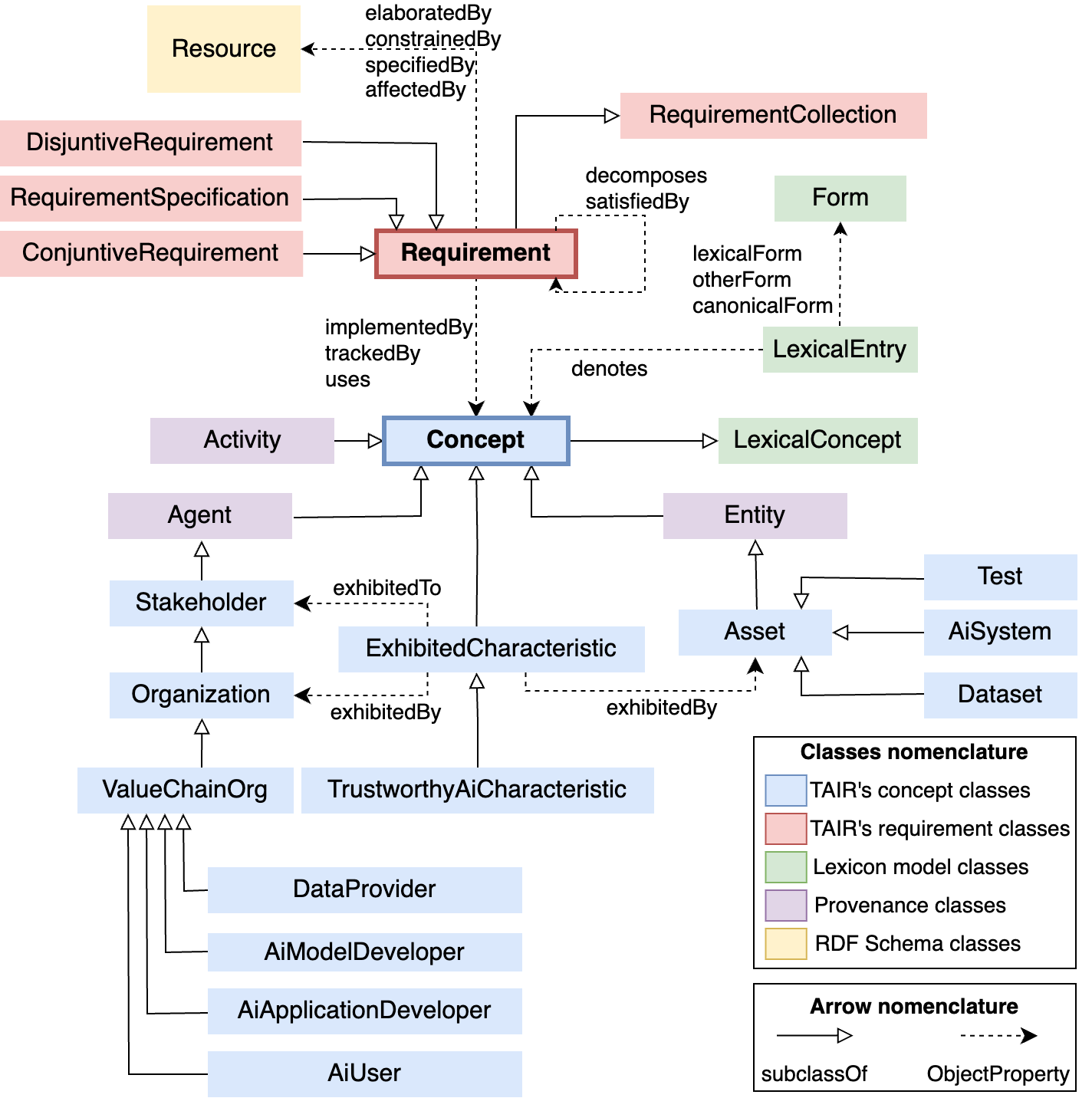}}
    \caption{Key concepts and relations in TAIR ontology}
    \label{fig:tair_ontology}
\end{figure}

The Trustworthy AI Requirements (TAIR) ontology\footnote{TAIR webpage: \url{https://tair.adaptcentre.ie/}} provides the elements to describe terms and requirements associated with a specific regulation or standard. \textbf{Figure~\ref{fig:tair_ontology}} depicts the TAIR ontology, where \texttt{Requirement} and \texttt{Concept} are the main classes in the ontology.

 The \texttt{Concept} class is a subclass of the \texttt{OntoLex}\footnote{\url{https://www.w3.org/2019/09/lexicog/}} vocabulary, which describes linguistic resources such as the representation of dictionaries or annotations commonly found in lexicography. 
 The \texttt{Requirement} class is used to describe normative clauses. A requirement could be related to a particular concept or lexical entry; this relationship is denoted by the properties \texttt{implementedBy} (who is responsible for implementing the described requirement), \texttt{trackedBy} (who tracks the updates of the requirement), and \texttt{uses} (who uses the described requirement).

\subsubsection{Requirements and Concepts Semantic Mappings}

The TAIR ontology aims to map regulations and standards requirements using linked data resources, making them available for consultation and query. Mapping requirements into linked data resources will help create systems capable of defining the requirements needed to comply with a domain-specific standard, such as information security and quality management. Additionally, it enables the identification and representation of concepts related to a standard, i.e., the words or phrases defined in the document with a specific meaning.  

\begin{figure}[!]
    %\sidecaption
    \centering
    \fbox{\includegraphics[scale=0.24]{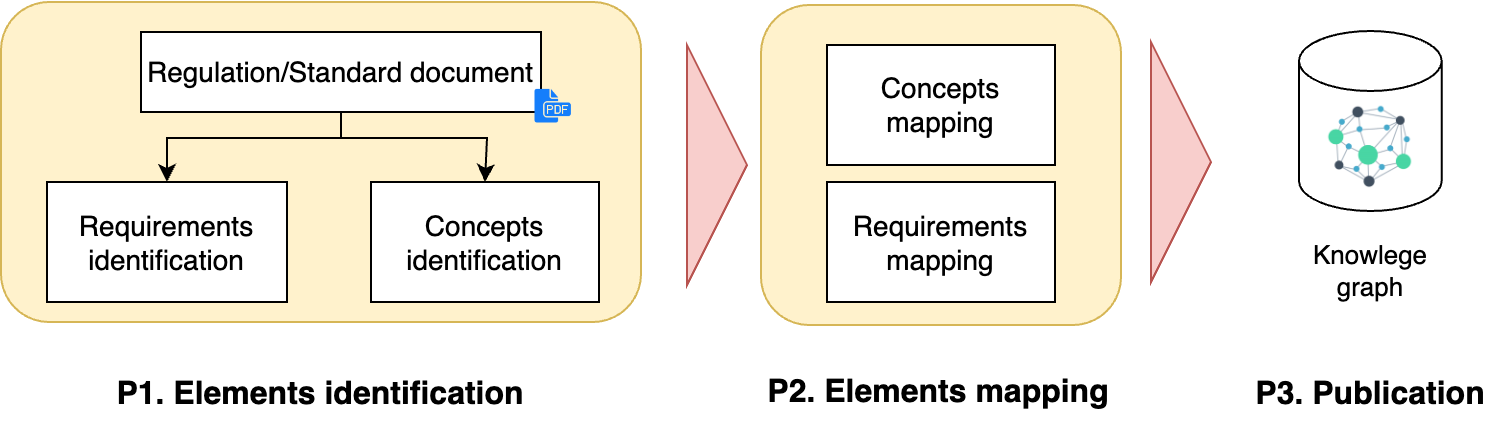}}
    \caption{The three phases in the regulation and standards mapping process}
    \label{fig:mapping_process}
\end{figure}

The mapping process (\textbf{Figure~\ref{fig:mapping_process}}) considers the regulation or standard document structure divided into clauses. The three phases (P1-P3) of the semantic mapping are described in the following paragraphs.

\paragraph{P1 - Elements identification} In this phase are identified the concepts and requirements for a regulation or standard. Concepts are usually defined in a section called ``Terms and definitions'' or  ``Definitions''. The requirements identification consists of looking for clauses expressed in the verbal form of shall or shall not~\footnote{ISO/IEC Directives, Part 2 - \url{https://www.iso.org/sites/directives/current/part2/index.xhtml}}. \textbf{Table~\ref{tab:ai_act_concepts}} exemplifies the type of concepts from the AI Act divided into actor (e.g., provider, user), artefact (e.g., AI System, Performance), and process (e.g., Putting into service, Withdrawal) concepts. %And, \textbf{Figure~\ref{fig:example_requirement}} presents an excerpt of a requirement clause in the verbal form of \emph{shall} from the harmonized structure for MSS.

\begin{table}[h]
\caption{\textit{Concepts defined in the AI Act.}}
\label{tab:ai_act_concepts}
\begin{tabular}{@{}lll@{}}
\toprule
\multicolumn{1}{c}{\textbf{Actor concepts}} &
  \multicolumn{1}{c}{\textbf{Artefact concepts}} &
  \multicolumn{1}{c}{\textbf{Process concepts}} \\ \midrule
\multicolumn{3}{l}{\cellcolor[HTML]{EFEFEF}\textbf{Concepts from AI Act definition (Article 3)}} \\ \midrule
\begin{tabular}[c]{@{}l@{}}Provider - \textbf{Small Scale} \\ \textbf{Provider} - User - \\ \textbf{Authorised Representative} -\\ Importer - \textbf{Distributor} -\\ Operator - \textbf{Notifying} \\ \textbf{Authority} - Conformity \\ Assessment Body - \textbf{Notified}\\ \textbf{Body} - Market Surveillance\\ Authority - \textbf{Law Enforcement}\\ \textbf{Authority} - \textbf{National} \\ \textbf{Supervisory Authority} -\\ National Competent \\ Authority.\end{tabular} &
  \begin{tabular}[c]{@{}l@{}}AI System - \textbf{Intended Purpose} - \\ Reasonably Foreseeable Misuse -\\ \textbf{Safety Component} - Instructions \\ For Use - \textbf{Performance} - \\ Substantial Change - \textbf{CE Marking} - \\ Harmonized Standard - \textbf{Common} \\ \textbf{Specification} - Training Data -\\ \textbf{Validation Data} - Testing Data - \\ \textbf{Input Data} - Biometric Data - \\ \textbf{Emotion Recognition System} - \\ Biometric Categorization System - \\ \textbf{Remote Biometric Identification} \\ \textbf{System} - Real-Time Remote \\ Biometric Identification System -\\ \textbf{Post Remote Biometric}.\end{tabular} &
  \begin{tabular}[c]{@{}l@{}}Placing On The Market - \\ \textbf{Making Available On The} \\ \textbf{Market} - Putting Into \\ Service - \textbf{Recall} - \\ Withdrawal - \textbf{Conformity} \\ \textbf{Assessment} - Post Market \\ Monitoring - \textbf{Law} \\ \textbf{Enforcement}.\end{tabular} \\
\bottomrule
\end{tabular}
\end{table}

%\begin{figure}[h]
    %\sidecaption
%    \centering
%    \includegraphics[scale=0.25]{Figures/Example-Requirement.png}
%    \caption{\textit{Example of requirement expressed in the verbal form of shall.}}
%    \label{fig:example_requirement}
%\end{figure}

\begin{figure}[h]
    %\sidecaption
    \centering
    \fbox{\includegraphics[scale=0.25]{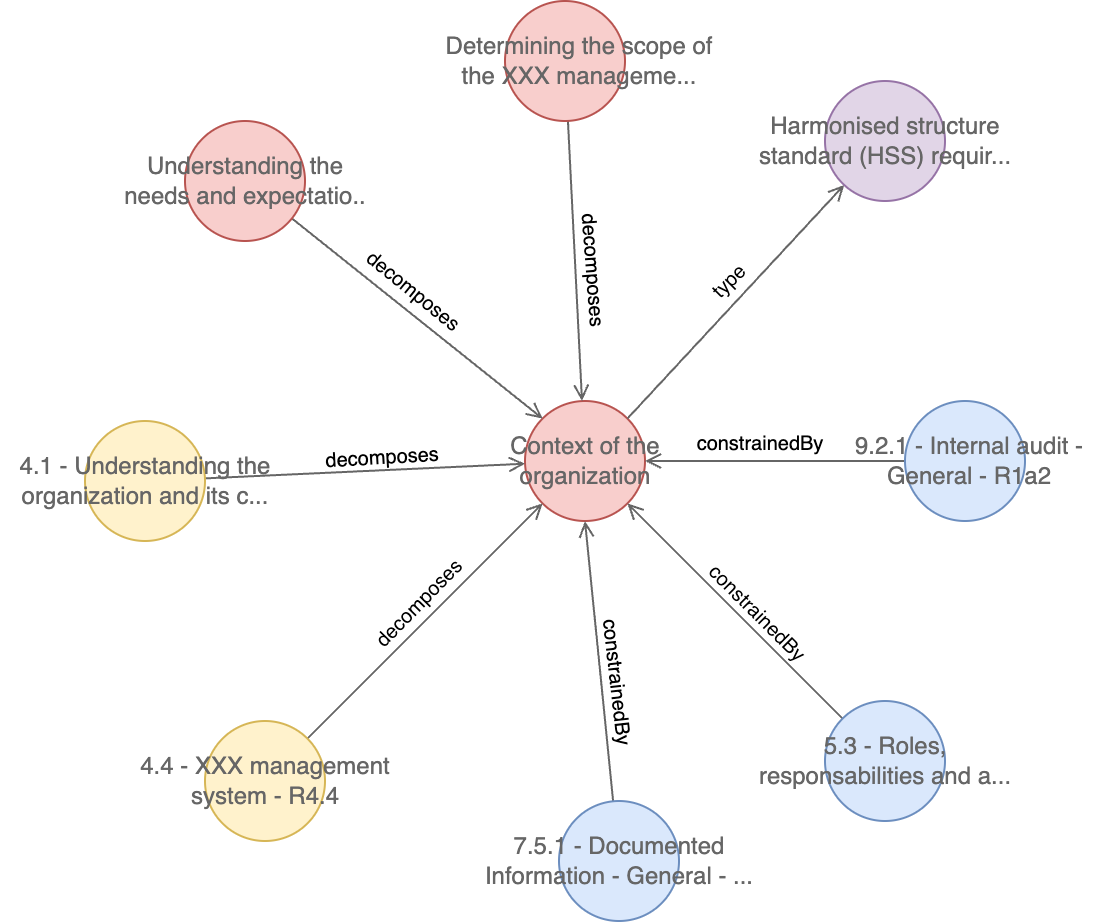}}
    \caption{\texttt{RequirementCollection} class example for the requirement collection ``Context of the organization'' extracted from the harmonized structure for MSS. The \texttt{decomposes} property defines the collection's requirements.}
    \label{fig:example_requirementCollection}
\end{figure}

\begin{figure}[h]
    %\sidecaption
    \centering
    \fbox{\includegraphics[scale=0.25]{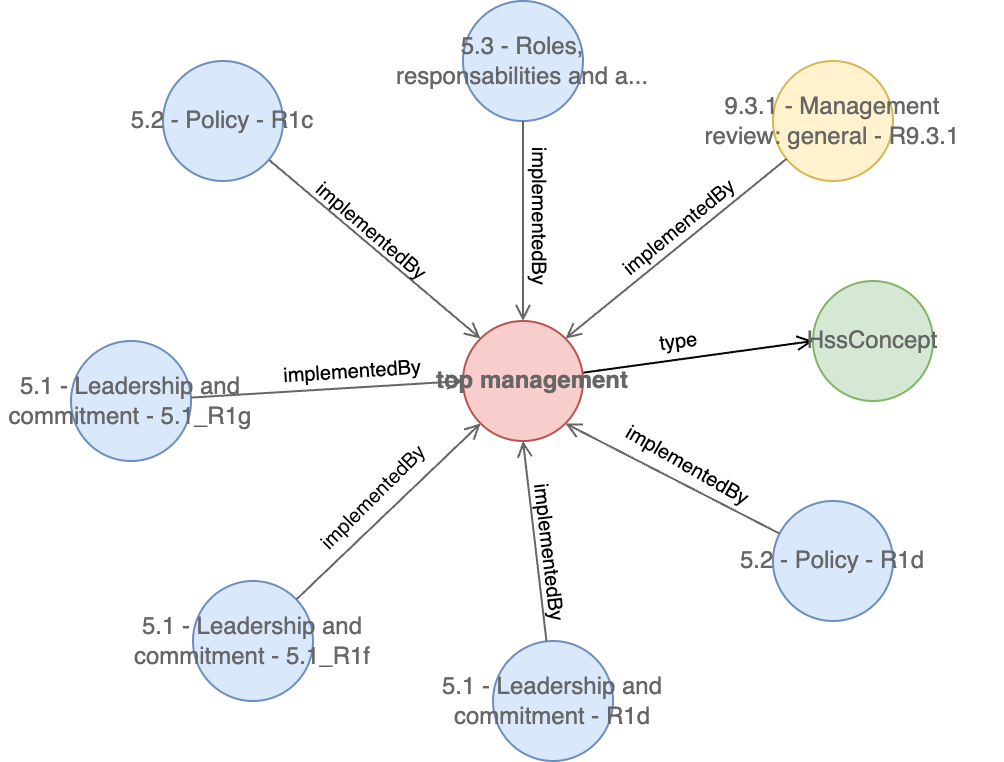}}
    \caption{Concept's mapping example though the ``Top management'' concept extracted from the harmonized structure MSS. The \texttt{implementedBy} property defines the relationship between a concept and a requirement.}
    \label{fig:example_concept}
\end{figure}

\paragraph{P2 - Elements mapping} This phase describes each requirement and concept definition into a linked data element considering the classes and properties of the TAIR ontology. The \texttt{RequirementCollection} class is used to define that a set of requirements belongs to the same clause or article, e.g., \textbf{Figure~\ref{fig:example_requirementCollection}} exemplifies the requirement collect ``Context of the organization'' from the harmonized structure for MSS, where \texttt{decomposes} property defines the requirements associated with the collection. The \texttt{Requirement} class describes a particular requirement from the clause or article. The \texttt{Concept} class describes a particular regulation or standard concept, e.g., \textbf{Figure~\ref{fig:example_concept}} the concept of ``top management'' from the harmonized structure for MSS. Finally, a concept is associated with a specific requirement by means of the properties \texttt{uses} and \texttt{implementedBy} if it is directly mentioned in the requirement.

\paragraph{P3 - Publication} Provides the mechanisms to access the ontology documentation and query the requirements and concepts. The Ontotext GraphDB\footnote{\url{https://graphdb.ontotext.com/}} graph database was used to publish the TAIR ontology. GraphDB is a triplestore with RDF and SPARQL support and graph visualization capabilities. 

\begin{figure}[h]
    %\sidecaption
    \centering
    \fbox{\includegraphics[scale=0.35]{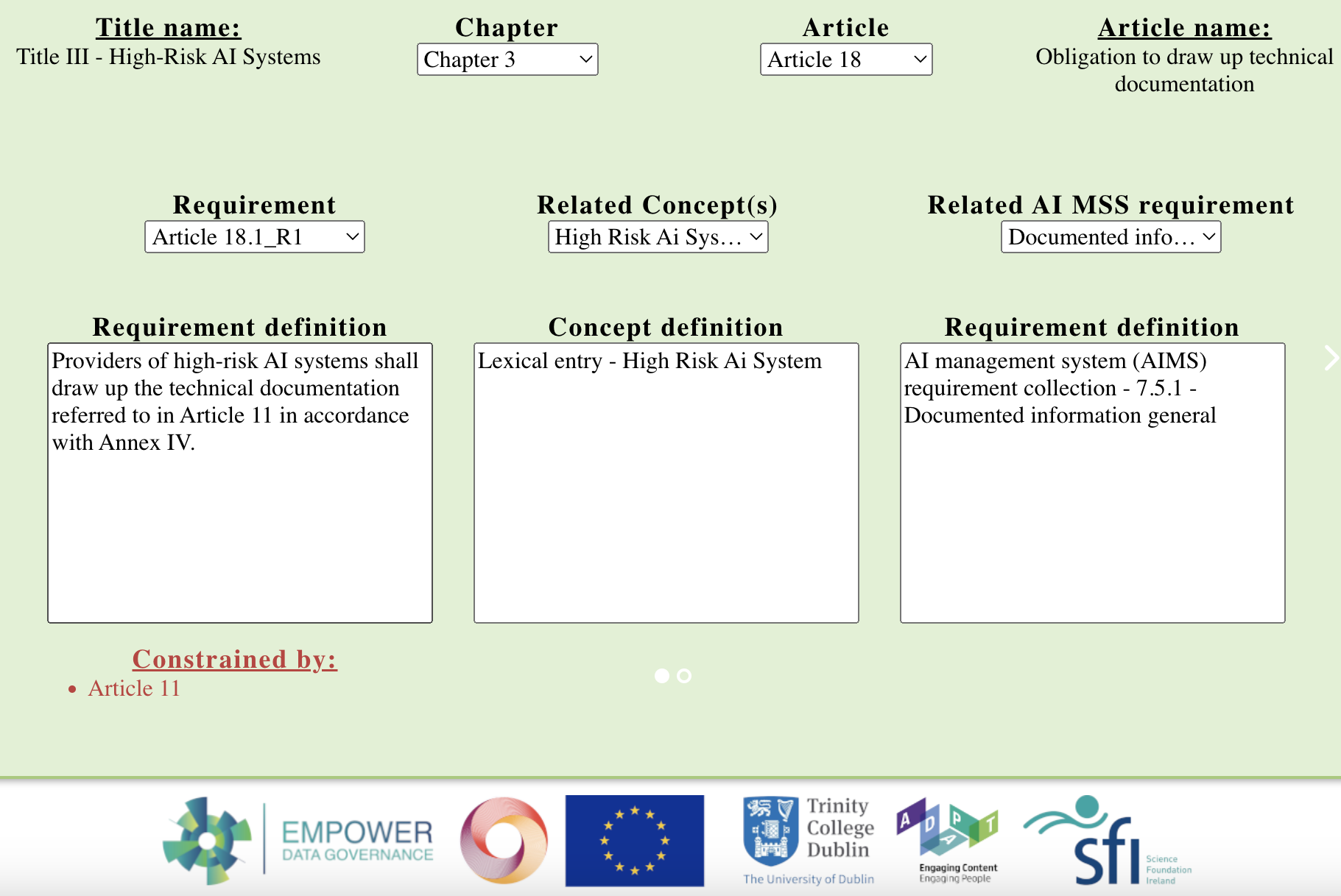}}
    \caption{TAIR ontology requirements demo. It exemplifies a requirement from article 18, identifying the concepts mentioned in the requirement and the related AI MSS requirement.}
    \label{fig:demo_requirements}
\end{figure}

\begin{figure}[h]
    %\sidecaption
    \centering
    \fbox{\includegraphics[scale=0.35]{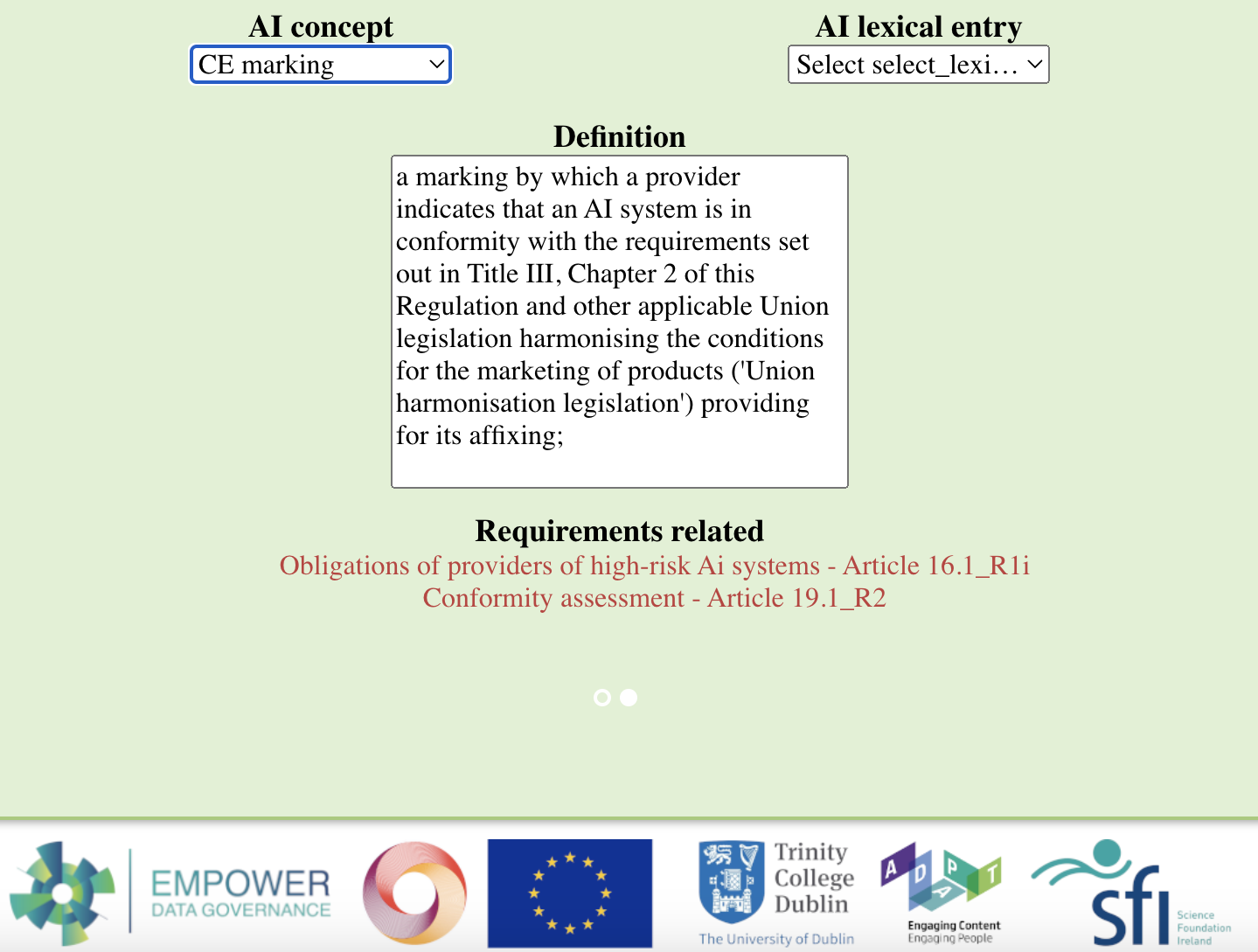}}
    \caption{TAIR ontology concepts demo. It exemplifies the definition of the CE marking concept and the requirements that are mentioned.}
    \label{fig:demo_concepts}
\end{figure}

Two demos~\footnote{\url{TAIR demo: https://tair.adaptcentre.ie/demo.html}} of the TAIR ontology were developed, focusing on the requirements and concepts of the Draf AI Act. The first demo (\textbf{Figure~\ref{fig:demo_requirements}}) explores Title III of the Draft AI Act related to High-Risk AI System requirements. The second demo (\textbf{Figure~\ref{fig:demo_concepts}})explores the concepts from the Draft AI Act.

The extraction of requirements from the AI Act related to compliance obligations on AI providers resulted in 118 separate requirements. Where relevant, these are linked to the 46 explicitly defined concepts from Article 3 (\textbf{Table~\ref{tab:ai_act_concepts}}) of the AI Act. Additionally, 23 lexical entries were extracted from the AI Act requirements. (\textbf{Table~\ref{tab:ai_act_lexical_entries}}).

\begin{table}[]
\caption{\textit{Lexical entries extracted from its compliance requirements statements.}}
\label{tab:ai_act_lexical_entries}
\begin{tabular}{@{}lll@{}}
\toprule
\multicolumn{1}{c}{\textbf{Actor concepts}} &
  \multicolumn{1}{c}{\textbf{Artefact concepts}} &
  \multicolumn{1}{c}{\textbf{Process concepts}} \\ \midrule
\multicolumn{3}{l}{\cellcolor[HTML]{EFEFEF}\textbf{\begin{tabular}[c]{@{}l@{}}Concepts from extraction of AI Provider compliance (Chapter 2, on requirements for \\ high-risk AI systems, Articles 8 to 13, and 15; and chapter 3, on obligations for AI \\ providers, Articles 16 to 23)\end{tabular}}} \\ \midrule
children &
  \begin{tabular}[c]{@{}l@{}}high risk AI system - \textbf{risk} \\ \textbf{management system} - risk \\ management measures -\\ \textbf{harmonized standards} - residual \\ risk - \textbf{preliminarily defined} \\ \textbf{metrics} - preliminarily defined \\ probabilistic thresholds - \\ \textbf{technical documentation} - logs -\\ \textbf{quality management system} - \\ quality assurance system - \textbf{risk}.\end{tabular} &
  \begin{tabular}[c]{@{}l@{}}continuous iterative process -\\ \textbf{eliminating or reducing risks} -\\ perform consistently for their \\ intended purpose - \textbf{testing of} \\ \textbf{high-risk AI systems} - testing \\ procedures - \textbf{development} \\ \textbf{process} - placing on the \\ market - \textbf{putting into service} -\\ children - \textbf{human oversight} -\\ quality control - \textbf{resource} \\ \textbf{management}.\end{tabular} \\ \bottomrule
\end{tabular}
\end{table}

%\del{\section{Discussion of Benefits and Applications}} \label{benefits}
%\del{mapping the AI Act requirements to standards, determining the sufficiency of standards for compliance with the AI Act, mapping legal requirements to trustworthy AI principles+integration of the AI Act with harmonization legislation (Annex II) --> terms and definitions such as \emph{safety component}    }

\subsection{Ontology evaluation} 

This evaluation considers ontology design best practices to detect errors or inconsistencies in the ontology structure, i.e., how the syntax of an ontology representation conforms to an ontology language~\cite{aruna2011survey}. 

The TAIR ontology language conformity evaluation was conducted through the OntOlogy Pitfall Scanner! (OOPS!) tool~\cite{poveda2014oops}. The OOPS! tool detects potential problems in the provided ontology by means of a semi-automatic diagnosis for 32 pitfalls. The evaluation result is classified as minor, important, and critical according to the pitfall detected. Each pitfall is associated with an importance level decided in conjunction with OOPS! developers, experienced ontological engineers, and users. For example, a pitfall classified as critical occurs if the ontology is not available (documentation not available online).

The OOPS! tool implements three pitfall detection methods: Structural pattern matching, Lexical content analysis, and Specific characteristic search. The former, which implements 24 of the 32 pitfalls, analyzes the internal structure of the ontology, looking for a particular structural pattern that spotted a pitfall. The lexical content analysis method, which implements 9 of the 32 pitfalls, analyzes lexical entities based on the content of annotations (e.g., \texttt{rdfs:label}) and identifiers for detecting pitfalls. The latter method, which implements 5 of the 32 pitfalls, checks for general characteristics of the ontology unrelated to previous methods, e.g., the name given to the ontology does not contain file extensions.

%The evaluation results are shown in \textbf{Table~\ref{tab:languageConformityEvaluation}}. 
The pitfalls identify by the OOPS! tool are minor problems, i.e., do not represent a problem. The most recurrent pitfall is the missing definition of inverse relationships, e.g., the inverse property \texttt{constrains} is not defined for the property \texttt{constrainedBy}. The missing annotation pitfalls refer to properties and/or classes without a human-readable property; it mainly occurs for external classes defined in the TAIR ontology, such as \texttt{LexicalConcept} or \texttt{Resource} classes. Finally, the unconnected ontology elements pitfall occurs because a defined class is not connected with any other element of the ontology, e.g., the class \texttt{LexicalConcept} is not connected with any other class; the class \texttt{Concept} refers to it but only as their subclass. All the unconnected ontology elements and missing annotation pitfalls reference external vocabularies, e.g., SKOS or RDFS; their definition will be found in the corresponding URL.  

\begin{comment}
\begin{table}[]
\caption{Language conformity evaluation}
\label{tab:languageConformityEvaluation}
\begin{tabular}{@{}clll@{}}
\toprule
\textbf{\# of cases} &
  \multicolumn{1}{c}{\textbf{Pitfall name}} &
  \multicolumn{1}{c}{\textbf{Pitfall description}} &
  \multicolumn{1}{c}{\textbf{Classification}} \\ \midrule
3 &
  \begin{tabular}[c]{@{}l@{}}Creating unconnected \\ ontology elements\end{tabular} &
  \begin{tabular}[c]{@{}l@{}}Ontology elements (classes, object properties and\\ datatype properties) are created isolated, with no\\ relation to the rest of the ontology.\end{tabular} &
  Minor \\ \midrule
6 &
  Missing annotations &
  \begin{tabular}[c]{@{}l@{}}This pitfall consists in creating an ontology element\\ and failing to provide human readable annotations\\ attached to it. Consequently, ontology elements lack\\ annotation properties that label them (e.g. rdfs:label, \\ lemon:LexicalEntry, skos:prefLabel or skos:altLabel)\\ or that define them (e.g. rdfs:comment or \\ dc:description).\end{tabular} &
  Minor \\ \midrule
18 &
  \begin{tabular}[c]{@{}l@{}}Inverse relationships not \\ explicitly declared\end{tabular} &
  \begin{tabular}[c]{@{}l@{}}This pitfall appears when any relationship (except \\ for those that are defined as symmetric properties \\ using owl:SymmetricProperty) does not have an \\ inverse relationship (owl:inverseOf) defined within \\ the ontology.\end{tabular} &
  Minor \\ \bottomrule
\end{tabular}
\end{table}

\end{comment}

\section{Conclusion and Future Work}\label{conclusion}

The Trustworthy AI Requirements (TAIR) ontology provides a basis for capturing and analyzing terms and requirements as concept sets from normative statements from the AI Act and the conformance-focused international standard on AI from SC42. This is made partially available as an Open Knowledge Graphs (OKG) resource that allows the links between defined terms, other relevant concepts, and the requirements themselves to be published in a traceable, queryable, and navigable manner. 

As this work was based on a draft version of the AI Act, we await the publication in early 2024 of the formal text in order to repeat the extraction of concepts and requirements and publish the second version of the TAIR ontology. We will then aim to promote this version of the model and its online exploration features to different potential groups who may find this useful to garner feedback on its utility. Such groups could include subject matter experts in specific high-risk application domains, such as healthcare or education, who may seek to build domain-specific extensions to concepts in this model. 

The model may be of use to policymakers and standards developers involved in the development of harmonized standards, in guidelines to support the implementation of the Act, such as EC guidelines to SME developing or public sector agencies procuring AI, and those establishing transparency mechanisms for regulatory learning mechanisms such as regulatory sandboxes and real-life trials. We would also seek feedback from scholars in law, ethics, social science, and information systems on whether this open approach to AI act concepts provides a basis for improving comparison, aggregation, and replication of studies in these areas. 

Further horizontal requirements mapping will be explored, especially as the SC42 AI Management System Standard (AI MSS) is supported by further standards, including ISO/IEC 23053:2022 (Framework for Artificial Intelligence (AI) Systems Using Machine Learning), ISO/IEC 23894:2023 (Guidance on risk management), ISO/IEC TR 24027:2021 (Bias in AI systems and AI aided decision making), ISO/IEC TR 24028:2020 (Overview of trustworthiness in artificial intelligence), ISO/IEC TR 24368:2022 (Overview of ethical and societal concerns), and ISO/IEC 38507:2022 (Governance implications of the use of artificial intelligence by organizations). However, fully realizing this potential would require agreement between the European Commission (EC)  and the European Standardization Organization (ESO) on how harmonized standards can be publicly available without the current paywall fees. 

In the long term, this approach and its open resources could be used to compare proprietary or national trustworthy AI mechanisms to the conformance and compliance system offered by the AI Act and its harmonized standards. The demand for such mappings is already apparent in the cross-walk  mappings\footnote{\url{https://www.nist.gov/itl/ai-risk-management-framework/crosswalks-nist-artificial-intelligence-risk-management-framework}} being developed by the US National Institute of Standards and Technology (NIST) between its AI Risk Management Framework and approaches present in ISO/IEC standards, AI Act and OECD models, albeit at a less fine-grained level of abstraction that demonstrated here. The more detailed mapping may also assist future policy alignment work by the EU-US Trade and Technology Council, which has already started to propose a common terminology and taxonomy between jurisdictions \footnote{\url{https://digital-strategy.ec.europa.eu/en/library/eu-us-terminology-and-taxonomy-artificial-intelligence}}. Finally, we hope such open mapping resources could also assist civil society organizations to monitor the future implementation and enforcement of the AI Act, especially in relation new arenas where regulation may be unclear or contested in relation to fundamental rights protections. 

\bmhead{Acknowledgements}

This project has received funding as a research gift from Meta and is supported by the Science Foundation Ireland under Grant Agreement No 13/RC/2106\_P2 at the ADAPT SFI Research Centre and the European Union’s Horizon 2020 Marie Skłodowska-Curie grant agreement No 813497 for the PROTECT ITN.

%\bibliography{bibliography}% common bib file


\end{document}